\documentclass[10pt, a4paper]{article}
\usepackage{amsmath}
\usepackage{tabularx,booktabs,adjustbox}
\usepackage{placeins}
\usepackage{pifont}
\usepackage{array}
\usepackage[most]{tcolorbox} 
\usepackage{xcolor}
\usepackage{tikz}
\usepackage[table,dvipsnames]{xcolor}

\usepackage{graphicx} 

\usepackage{makecell}
\newcommand{\cmark}{\textcolor{green}{\textbf{O}}}
\newcommand{\xmark}{\textcolor{red}{\textbf{X}}}
\usepackage[final]{lrec2026} 

\usepackage{multirow}

\title{Automatic Inter-document Multi-hop Scientific QA Generation}

\name{
Seungmin Lee$^{1,2}$, Dongha Kim$^{1,2}$, Yuni Jeon$^{1,2}$, 
Junyoung Koh$^{1}$, and Min Song\thanks{*Corresponding author}$^{1,2*}$}

\address{
$^{1}$Yonsei University, Seoul, Republic of Korea\\
$^{2}$OnomaAI, Seoul, Republic of Korea\\
\{elplaguister, eastha, jeonuni, solbon1212, min.song\}@yonsei.ac.kr
}

\abstract{
Existing automatic scientific question generation  studies mainly focus on single-document factoid QA, overlooking the inter-document reasoning crucial for scientific understanding. We present AIM-SciQA, an automated framework for generating multi-document, multi-hop scientific QA datasets. AIM-SciQA extracts single-hop QAs using large language models (LLMs) with machine reading comprehension and constructs cross-document relations based on embedding-based semantic alignment while selectively leveraging citation information. Applied to 8,211 PubMed Central papers, it produced 411,409 single-hop and 13,672 multi-hop QAs, forming the IM-SciQA dataset. Human and automatic validation confirmed high factual consistency, and experimental results demonstrate that IM-SciQA effectively differentiates reasoning capabilities across retrieval and QA stages, providing a realistic and interpretable benchmark for retrieval-augmented scientific reasoning. We further extend this framework to construct CIM-SciQA, a citation-guided variant achieving comparable performance to the Oracle setting, reinforcing the dataset’s validity and generality.
 \\ \newline  }

\begin{document}

\maketitleabstract

\section{Introduction}
Large language models (LLMs) play a pivotal role in scientific research, significantly advancing from simple text completion tools to sophisticated systems capable of complex reasoning and autonomous problem-solving. This progress has enabled "co-scientist" to independently generate hypotheses, conduct experiments, and write papers \cite{lu2024ai, gottweis2025towards}. LLMs now achieve near-human performance on question-answering (QA) benchmarks, and retrieval-augmented generation (RAG) facilitates sophisticated multi-document reasoning \cite{lewis2020retrieval}.

Despite these advances, multi-hop scientific QA remains a crucial challenge for enabling machines to reason across interconnected documents \cite{liu2023lost, wadden2022scifact, shi2024generate, dasigi2021qasper, park2025dochop}. This difficulty arises from the vast volume of literature, domain-specific terminology, and fragmented information distribution, all of which make cross-document reasoning inherently complex. Moreover, existing datasets are scarce and limited in scope: most benchmarks focus on single-document QA, failing to reflect the complexity of real scientific inquiries that span multiple sources. Although recent datasets such as M3SciQA \citep{li2024m3sciqa} have begun addressing multi-document scenarios, the landscape remains sparse, particularly for multi-hop QA where answers must be composed from multiple pieces of evidence. This data scarcity continues to hinder the development and evaluation of robust scientific QA models.

\begin{figure}
    \centering
    \includegraphics[width=\columnwidth]{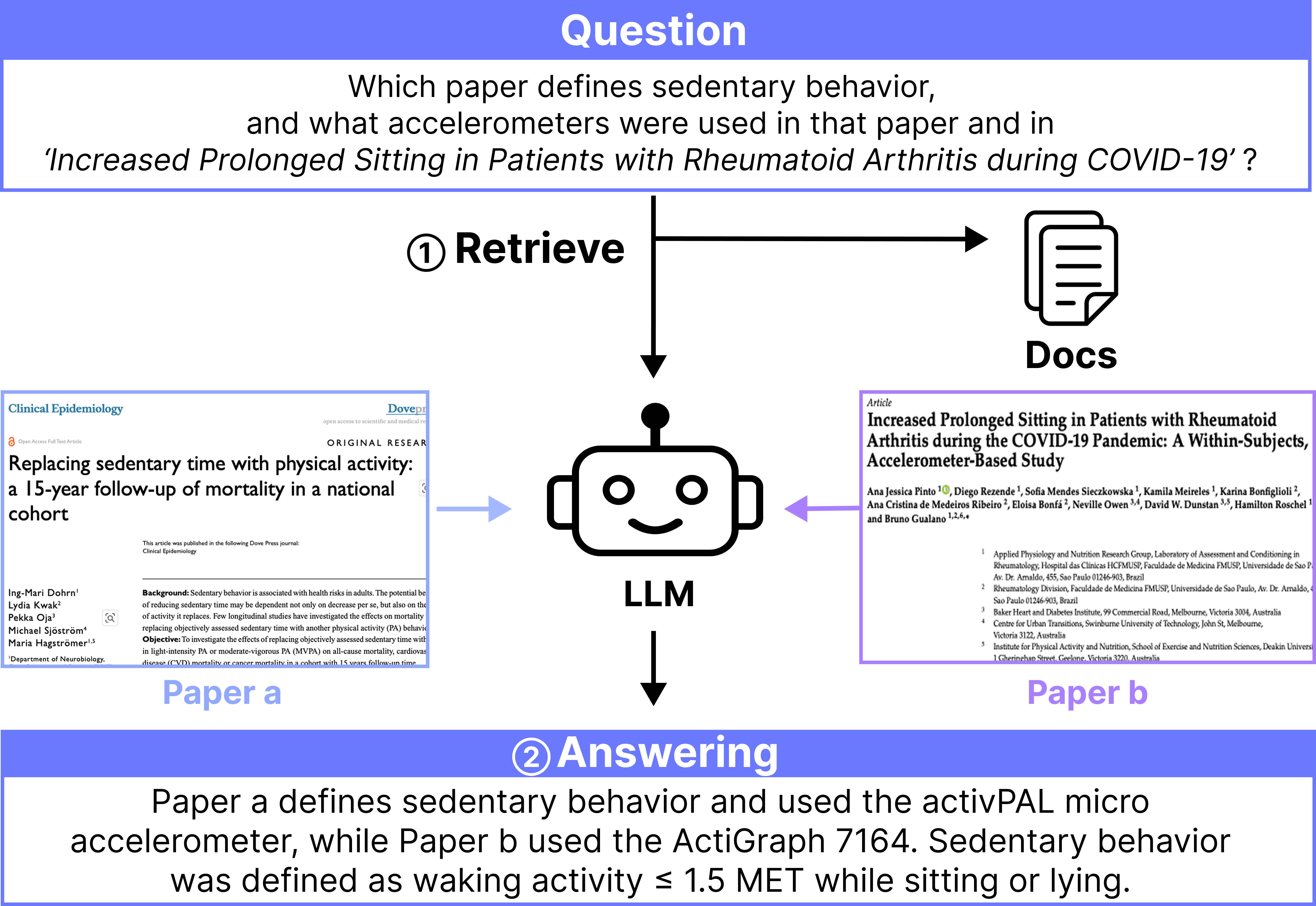}
    \caption{Example of Inter-document Multi-hop QA: \ding{172} the model retrieves relevant documents and \ding{173} integrates complementary evidence to answer.}
    \label{fig:fig1}
\end{figure}

\begin{table*}[!t]
\centering
\caption{Comparison with other datasets. \textbf{M hop}: Multi Hop, \textbf{M Doc}: Multi Documents, \textbf{Q Gen}: Question Generation method, \textbf{Free-form}: answers are open-ended.}
\label{tab:dataset_comparison}
\resizebox{\textwidth}{!}{
\begin{tabular}{lcccccll}
\toprule
\textbf{Dataset} & \textbf{M Hop} & \textbf{M Doc} & \textbf{\#QA} & \textbf{QA Gen} & \textbf{Free-form} & \textbf{Domain} & \textbf{Source} \\
\midrule
PubMedQA   & \xmark & \xmark & 211k & Human & \xmark & Biomedical & PubMed \\
SciQAG-24D     & \xmark & \xmark & 180k & LLM & \cmark & Scientific Literature & Scientific Literature \\
WikiHop    & \cmark & \cmark & 51k & Rule-based & \xmark & General/Wiki & Wiki \\
HotpotQA   & \cmark & \cmark & 113k & Human & \cmark & General/Wiki & Wiki \\
DocHop-QA  & \cmark & \cmark & 11k & LLM & \xmark & Biomedical/Scientific & PubMed \\
M3SciQA    & \cmark & \cmark & 1.4k & Human & \xmark & NLP Papers & NLP Papers (+cited) \\
\midrule
\textbf{IM-SciQA (ours)} & \cmark & \cmark & 13.4K & LLM & \cmark & \textbf{Biomedical} & \textbf{PubMed} \\
\bottomrule
\end{tabular}
}
\end{table*}

A key bottleneck in scientific QA is curating datasets that are both diverse and challenging.
Manual annotation of scientific documents is costly and requires domain experts.
For instance, PubMedQA \citep{jin2019pubmedqa} combined 1,000 expert-labeled items with over 210k automatically generated instances, underscoring the scalability of automatic question generation (QG).
However, most automated QG efforts focus on general-domain sources such as Wikipedia, with little attention to primary scientific documents.
Scientific documents present additional challenges, including specialized terminology, dense exposition, explicit citation networks, and implicit relations across papers.
Effective QG must explicitly leverage or robustly accommodate these characteristics.

In this paper, we propose Automatic Inter-document Multi-hop Scientific QA generation (AIM-SciQA), an automated QG methodology tailored for scientific QA.
Our approach uses a specialized LLM in a machine reading comprehension (MRC) paradigm to extract potential QA pairs from individual papers, then performs cross-paper matching to construct multi-document questions by aggregating information across documents.
\textbf{AIM-SciQA} considers both explicit citation links and implicit connections, enabling automatic generation of complex multi-document multi-hop QA.
We apply this pipeline to the biomedical domain, generating approximately 411,409, single-hop QA pairs from individual papers and deriving tens of thousands of multi-document, multi-hop examples.

To our knowledge, this is the first large-scale scientific QA dataset with automatically generated multi-hop questions spanning multiple documents in a specialized domain. Extensive experiments on the IM-SciQA and CIM-SciQA datasets demonstrate that AIM-SciQA produces high-quality multi-hop QA pairs and provides a realistic benchmark for evaluating retrieval and reasoning performance of large language models. 
With AIM-SciQA, we aim to advance the state of Scientific QA by enabling efficient dataset creation and providing a new benchmark to drive research in complex multi-document question answering on scientific texts. 
The datasets \href{https://huggingface.co/datasets/MRAGDATASET/Pubmed-MHQA}{IM-SciQA} and \href{https://huggingface.co/datasets/MRAGDATASET/Pubmed-Cited-MHQA}{CIM-SciQA} are publicly available on Hugging Face.

In summary, our contributions are as follows:

\begin{itemize}
    \item \textbf{AIM-SciQA Framework}: We introduce a novel automated QA Generation pipeline that generates inter-document, multi-hop QA dataset from scientific literature.
    \item \textbf{IM-SciQA Dataset}: By leveraging AIM-SciQA framework, we automatically created IM-SciQA, large-scale scientific inter-document multi-hop QA Dataset including 13,672 QAs from PubMed Central.
    \item \textbf{Comprehensive Analyses}: We conducted comprehensive evaluations of recent Encoders and LLMs on IM-SciQA, revealing the effectiveness of our restricted retrieval setting and completeness of inter-document QA.
\end{itemize}
\section{Related Works}

\begin{figure*}
    \centering
    \includegraphics[width=\textwidth]{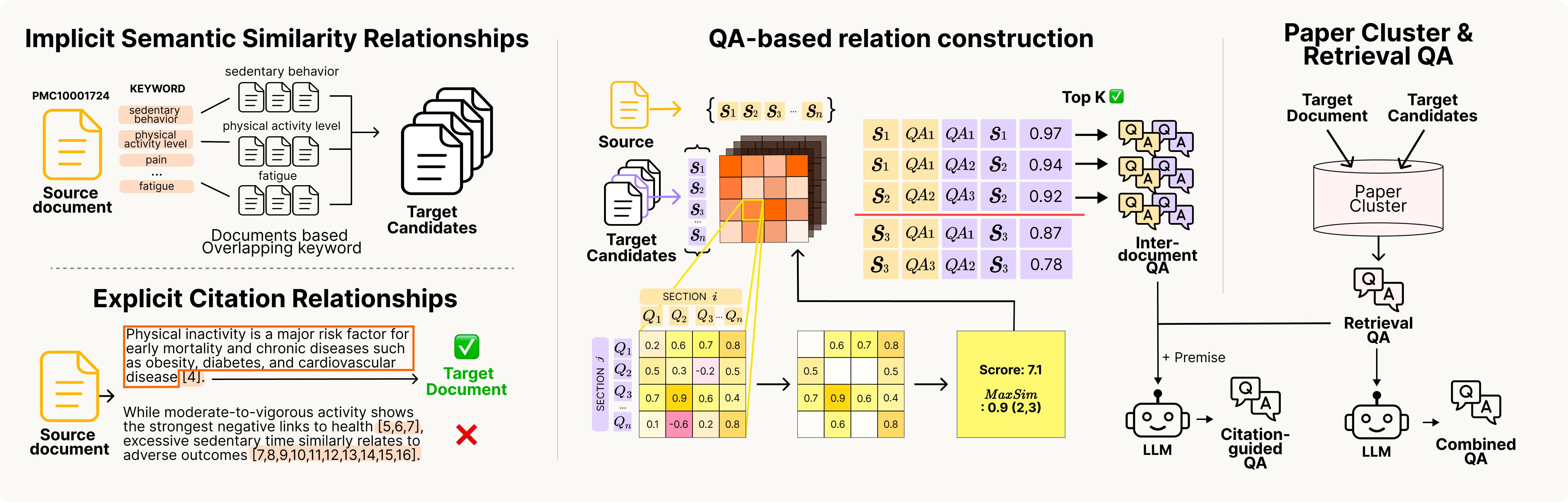}
    \caption{Illustration of the relation construction process in AIM-SciQA. For IM-SciQA, candidate relations are created from papers sharing overlapping keywords with the source paper (\textbf{implicit semantic similarity relationships}), while CIM-SciQA constructs relations directly from explicit citation links (\textbf{explicit citation relationships}). From these candidate relations, \textbf{QA-based relation construction} (Section \ref{sec:document_relation}) process selects promising single-hop QA (made by Section \ref{sec:shqg}) pairs across papers to form inter-document QA candidates, and a \textbf{paper cluster} is built to define the target paper and its corresponding \textbf{retrieval QA}.}
    \label{fig:fig2}
\end{figure*}
\subsection{Scientific QA Datasets}
Early scientific QA centered on single-document understanding as seen in PubMedQA \citep{jin2019pubmedqa} (abstract-level, yes/no/maybe), COVID-QA \citep{moller2020covid} (expert-curated, topic-specific), ScienceQA \citep{lu2022scienceqa} (multimodal, MCQ), and QASPER \citep{dasigi2021qasper} (full-paper comprehension).
Domain-focused resources such as QASA \citep{lee2023qasa}, SPIQA \citep{pramanick2024spiqa} and BioASQ \citep{krithara2023bioasq} expanded coverage but largely retained single-document or constrained formats.
Multi-hop and cross-document settings emerged with M3SciQA \citep{li2024m3sciqa}, and domain extensions (e.g., ContractNLI \citep{koreeda2021contractnli}, ChemRxivQuest \citep{amiri2025chemrxivquest}). 
Recent biomedical efforts—DocHop-QA \citep{park2025dochop} and PeerQA \citep{baumgartner2025peerqa} advance realism and scale but often rely on predefined templates or limited question types. 

Overall, existing benchmarks as shown in Table \ref{tab:dataset_comparison}, either confine reasoning to a single document or constrain question diversity, motivating free-form, multi-document, multi-hop QA benchmark.

\subsection{Automatic Question Generation}
To reduce annotation cost and increase realism, automated QG has gained traction. 
LIQUID \cite{lee2023liquid} scales list-style QG on wikipedia but not multi-hop.
SciQAG \cite{wan2024sciqag} shows LLM-based QG at scale across scientific domains while remaining primarily single-document.
Multi-hop QG without rigid templates has been explored via end-to-end question rewriting (E2EQR) \cite{hwang2024e2eqr} and type-aware CoT with semantic extraction (TASE-CoT) \cite{lin2024tase}.
DocHop-QA applies LLMs to biomedical multi-document QG yet guides generation with 11 predefined concepts, potentially limiting naturalness. 

Our framework focuses on automatically constructing inter-document, multi-hop QA for rigorous evaluation of inter-document synthesis.
\section{Problem Statement}
Inter-document Multi-hop QA (IM-QA) aims to answer a question by integrating information distributed across multiple documents, as illustrated in Figure~\ref{fig:fig1}. Let $D$ denote the collection of all documents, $Q$ the set of questions, and $A$ the set of possible answers. For a given question $q \in Q$, the goal is to identify a subset of documents $D_q \subseteq D$ and to derive an answer $a \in A$ from them. Formally, this task can be expressed as follows:

\begin{equation}
a = f(q, D_q), \quad \text{where } |D_q| > 1.
\end{equation}

$f$ denotes the reasoning function that derives an answer $a$ by integrating evidence from multiple relevant documents $D_q$. This process can be divided into two stages— (1) \textbf{retrieving} a subset of relevant documents $D_q$ from the entire collection $D$ and (2) \textbf{synthesizing} the aggregated information within $D_q$ to produce the final answer.
We adopt a source-provided IM-QA setting, where each question is associated with a predefined set of source documents $S_q$.
This constraint enables controlled and interpretable retrieval evaluation while retaining the core aspect of multi-document reasoning.
\section{AIM-SciQA}
\label{sec:aim-sciqa}

We propose \textbf{AIM-SciQA}, an automated framework for constructing IM-QA over scientific documents.
We additionally build \textbf{IM-SciQA}, a biomedical paper–based IM-QA dataset.
The following section describes how IM-SciQA is constructed and concurrently provides a detailed account of the AIM-SciQA.
Our framework comprises as follows:
(1) paper selection, (2) \textbf{single-hop QA (SHQA) generation}, (3) \textbf{document relation construction}, and (4) \textbf{multi-hop QA (MHQA) generation}.

\subsection{Paper Selection}
\label{sec:paper_selection}
We collected 20,281 biomedical research articles available in PubMed Central (PMC) \footnote{https://pmc.ncbi.nlm.nih.gov/} between 2019 and 2024 to construct the initial corpus. We then applied filtering criteria to ensure that each paper provides complete metadata, including full text, abstract, and reference information. Because cited papers are not always available within the PMC subset, we restricted relation construction to citations pointing to papers that are also accessible in PMC and retained only papers with at least three such accessible citations. After applying these filters, 8,211 papers were retained as valid source documents for dataset construction.

\subsection{Single Hop QA Generation}
\label{sec:shqg}
Then, we automatically generated single-hop QA pairs that can be answered within a single paper.
This step provides the atomic units for constructing multi-hop QA by enabling effective combination of information across papers.
To identify important information in each paper and extract it in a QA format, we utilized \textit{Qwen2.5-7B-Instruct} \cite{qwen2.5}.
Specifically, given a document $D$ and one of its sections $S$, an instruction-tuned $LLM_\theta$ is prompted to produce a question-answer-evidence triplet:
\begin{equation}
\begin{split}
    LLM_\theta(D, S) & \longrightarrow (Q, A, E)
\end{split}
\end{equation}
where $Q$ asks for an information slot, $A$ provides its value, and $E$ is the supporting sentence in the text.

\begin{figure}[t]
    \centering
    \includegraphics[width=\columnwidth]{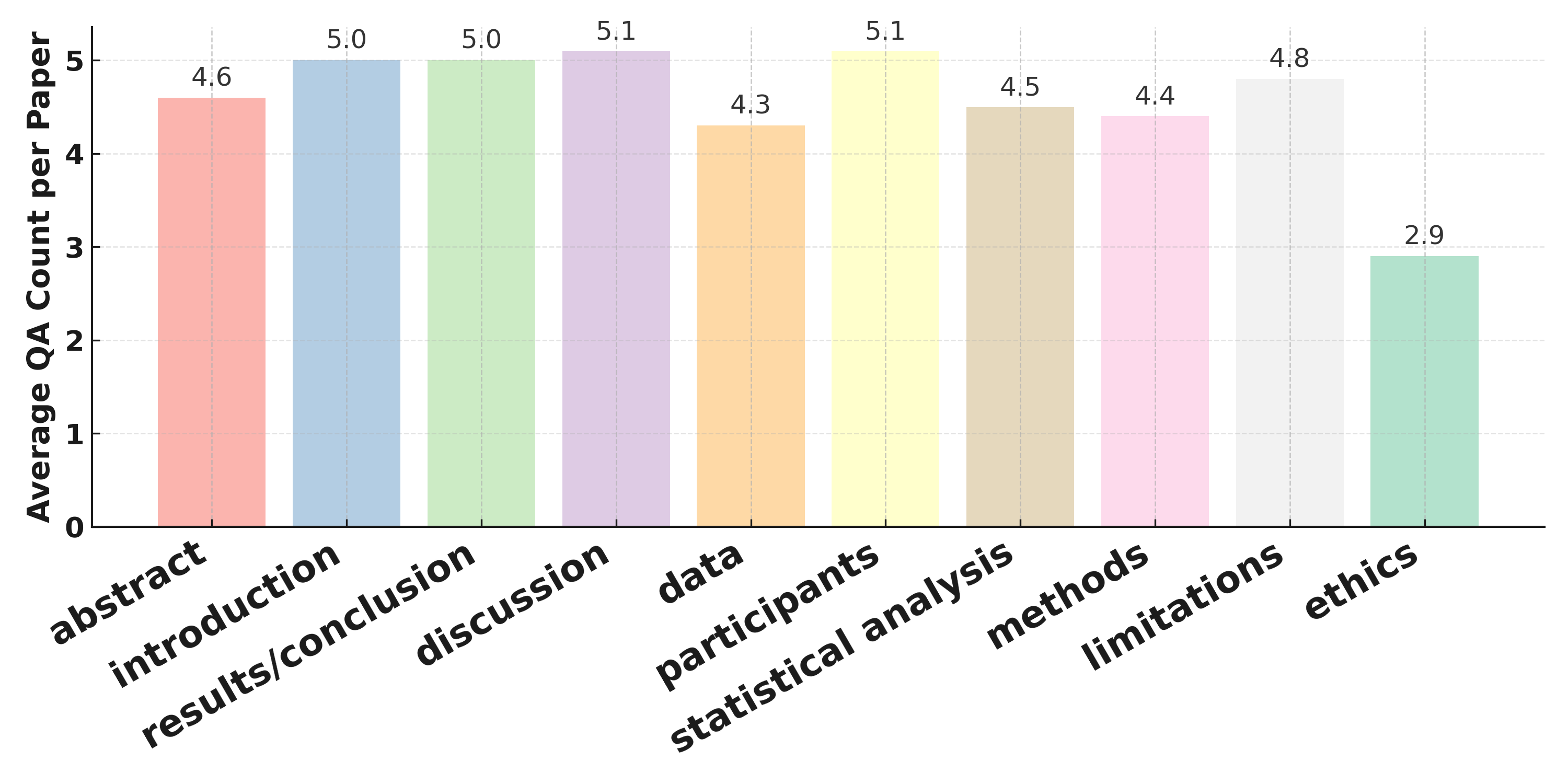}
    \caption{Average Single-hop QA Count by Section}
    \label{fig:fig_avg_QA_pub}
\end{figure}

\begin{figure}[t]
    \centering
    \includegraphics[width=\columnwidth]{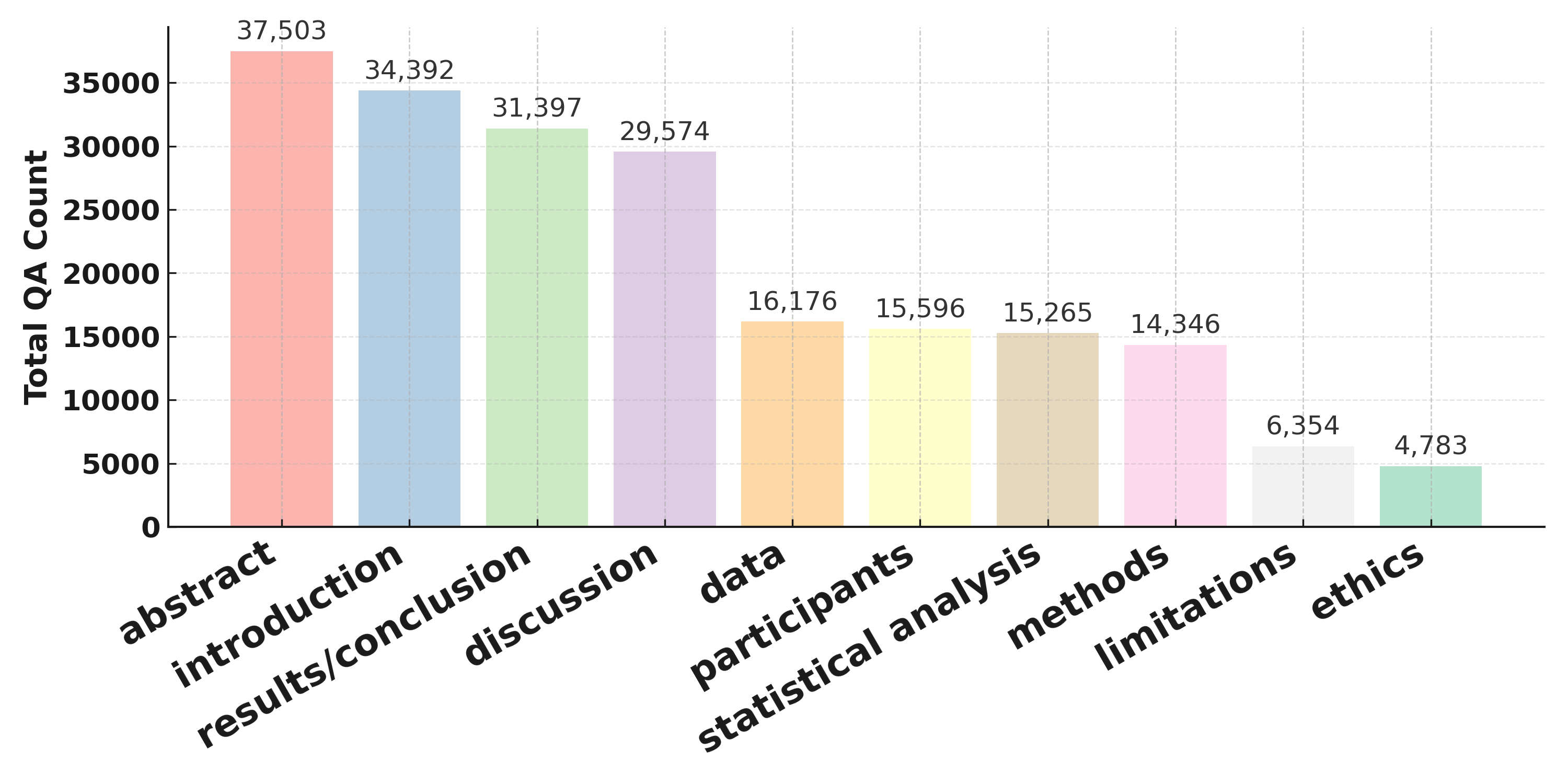}
    \caption{Total Single-hop QA Count by Section}
    \label{fig:fig_total_QA_pub}
\end{figure}

We designed a prompt to generate the list of triplets from each paragraph, where the evidence is the original sentence that the model referenced when creating the answer.
The prompt used for SHQA generation is provided in Appendix~\ref{app:shqa_prompt}.
To ensure quality, we only retained QA pairs where the evidence sentence exactly matched the paper's text, eliminating hallucinations.
This process yielded 457K SHQA pairs; their statistics are summarized in Figure \ref{fig:fig_avg_QA_pub}, Figure \ref{fig:fig_total_QA_pub}, and Table \ref{tab:qa_statistics}.

\paragraph{Quality Assurance}
Quality verification was conducted in two complementary phases.
First, a manual evaluation was performed by four graduate students on 210 randomly sampled QA pairs, assessing appropriateness on a 0–0.5–1 scale. 
Among them, 202 pairs were rated perfect, achieving 0.94 Gwet’s AC1 score, which corresponds to almost perfect agreement levels \cite{gwet2008computing,landis1977measurement}.
Closer inspection of the remaining error cases revealed that most errors occurred when the questions and evidence were not semantically aligned.
Notably, these mismatched cases showed an average $Sim(Q,A) - Sim(Q,E)$ approximately 0.1 lower than that of well-matched pairs.
Based on this observation, we implemented an automatic filtering procedure that removed the lowest 10\% of QA pairs by this similarity difference, effectively discarding samples most likely to contain misaligned evidence–answer mappings.
After this two-stage verification, a total of 411,409 SHQA pairs were retained as the final SHQA dataset.

\begin{table}[t]
\centering
\caption{Statistics of the Single-hop QA}
\label{tab:qa_statistics}
\begin{tabular}{l c}
\toprule
\textbf{Item} & \textbf{Value} \\
\midrule
Count & 411{,}409 \\
\midrule
Avg. Question Length & 88.75 \\
Avg. Answer Length & 126.92 \\
Avg. QA per Paper & 55.67 \\
Avg. QA per Section & 10.49 \\
\bottomrule
\end{tabular}
\end{table}

\subsection{Document Relation Construction}
\label{sec:document_relation}
We constructed inter-document relations for multi-hop QA through a unified pipeline primarily based on \textbf{QA-based semantic matching}, with an optional \textbf{Citation-guidance} as depicted in Figure~\ref{fig:fig2}.
Each document is represented as a collection of SHQA triplets $(Q, A, E)$, which act as proxies for section-level information units.
These triplets are encoded at the $Q$, $A$, and $QA$ levels for our PMC-derived SHQA corpus using the \textit{MedEmbed-large-v0.1} encoder \cite{balachandran2024medembed}, a model trained on the PMC-based BioASQ-QA dataset.
We represent each section by its question embeddings, not QA or answer embeddings. This choice is motivated by our goal: not to find sections with identical facts but to locate those whose information property or semantic attribute required by the question is similar. In other words, question embeddings enable us to align sections based on the similarity of their underlying informational attributes, rather than overly trivial content.

Formally, each document $D_i$ is defined as the set of its sections:
\begin{equation}
D_i = \{S_1^{(i)}, S_2^{(i)}, \ldots, S_x^{(i)}\}.
\end{equation}
Each section contains multiple embedded Question units.
For sections $S_i^{(1)}$ and $S_j^{(2)}$, let $\mathbf{u}_{i,p}$ denote the embedding of the $p$-th Q in $S_i^{(1)}$ and $\mathbf{v}_{j,q}$ that of the $q$-th Q in $S_j^{(2)}$.
We define the pairwise similarity matrix as:
\begin{equation}
K_{i,j}[p,q] = \operatorname{CosSim}(\mathbf{u}_{i,p}, \mathbf{v}_{j,q}).
\end{equation}
The section-level similarity is then computed as a weighted sum over pairs whose similarity exceeds the threshold $\tau = 0.3$:
\begin{equation}
\operatorname{Sim}(S_i^{(1)}, S_j^{(2)}) =
\sum_{p=1}^{n_i}\sum_{q=1}^{m_j}
\mathbf{1}(K_{i,j}[p,q] \ge \tau)
\cdot |K_{i,j}[p,q]|.
\end{equation}
This threshold suppresses irrelevant matches while preserving meaningful QA-level alignments and can be adjusted depending on the dataset characteristics or the embedding model used.
For each document pair $(D_1, D_2)$, we compute $\operatorname{Sim}$ across all section pairs and select the \textbf{Top-$K$ section pairs} with the highest similarity scores.
For each selected section pair, we identify the QA pair with the highest similarity score and use this score to represent the strength of the relation between the corresponding sections of the two documents.

Finally, we enforce diversity via top-k selection on mapped scores, keeping at most three IM-QA instances per document in the final set. As a result, we collected 14,365 candidate IM-QA items.

\paragraph{Paper Cluster Construction}
For each source $D_1$, we form a compact multiple-choice cluster of exactly 30 documents comprising the target document and up to 29 distractors. We start from the \textbf{keyword-overlap list} that consists of papers sharing keywords with $D_1$, capped at 29 and, if shorter, pad with random non-candidates without duplication. This cluster serves two purposes: (i) defining the answer set for first-hop target identification, and (ii) providing a contrast set for selecting a target-side retrieval QA whose content is most distinctive relative to the cluster.

\paragraph{Retrieval QA Selection}
To select a retrieval QA that uniquely identifies the target paper within the paper cluster, we aggregate all QA units of the target paper together with all QA units from the paper cluster.
Then, we compute QA embeddings for each item and calculate pairwise cosine distances across the entire pool.
For each QA in the target paper, we compute its distinctiveness by aggregating cosine distances to all QAs from other papers in the cluster, and select the QA with the highest total distance.
This QA is designated as the retrieval QA, as it remains grounded in the target paper while being most distinct from semantically similar candidates.

\paragraph{Citation-guided Relation Construction}
The citation-guided relation construction process follows the same procedure as above but replaces the target paper selection step and paper cluster construction step with citation evidence and cited paper pool, respectively.
For target paper selection step, instead of forming candidates through keyword overlaps, we enumerate all in-text citations in the source document $D_1$ and treat sentences containing a \textit{single citation marker} as grounded links to their referenced document $D_2$.
Formally, if $\mathcal{V}$ denotes the document set and $\mathcal{E}\subseteq\mathcal{V}\times\mathcal{V}$ represents directed citation edges, each sentence $s^\star$ in $D_1$ containing one reference marker deterministically maps to its bibliographic target $D_2$.
The paper cluster construction step is also replaced to the list of all cited paper that available publicly.
Thus, the citation-based variant retains the same embedding alignment and selection process as the QA-based approach, while substituting the candidate generation mechanism with citation-derived pairs.
To simulate and validate this procedure, we additionally developed Citation-guided Inter-document Multi-hop Scientific QA dataset (\textbf{CIM-SciQA}) with 74 items, built upon curated 51 papers which have all cited papers within our collected 8211 papers.

\subsection{Multi-hop Question Generation}
\label{sec:mhqg}
AIM-SciQA finally constructs multi-document, multi-hop questions from section-anchored $(Q,A,E)$ tuples in the source and target document.
Instead of simply splicing two QAs, we construct an explicit multi-step reasoning chain that integrates knowledge across papers.
Prior to generation, a pre-validation filters out non-relational pairs (e.g., lacking comparison, causation, or inference), resulting in 13,672 valid instances out of 14,365 candidates with 693 pre-rejected.
The remaining pairs then undergo a stepwise process to produce the final IM-QA.
Table \ref{tab:mhqa_statistics} summarizes the statistics of the resulting multi-hop QA dataset.
The full prompting template for question generation is provided in Appendix~\ref{app:mhqa_prompt}.

\begin{table}[t]
\centering
\caption{Statistics of the Multi-hop QA}
\label{tab:mhqa_statistics}
\begin{tabular}{lccc}
\toprule
\textbf{Item} & \textbf{Dev} & \textbf{Test} \\
\midrule
Count & 13,372 & 300 \\
\midrule
Avg. Retrieval Q Length & 15.05 & 15.16 \\
Avg. Inter-doc Q Length & 28.68 & 28.65 \\
Avg. Combined Q Length & 48.52 & 49.21 \\
\midrule
Avg. Inter-doc A Length & 28.13 & 29.98 \\
Avg. Combined A Length & 43.00 & 45.33 \\
\bottomrule
\end{tabular}
\end{table}

\paragraph{Step 1. Find Target Paper} We first reconstruct the SHQA associated with the target paper into a paper identification question, whose unique answer is the target document itself.
Consequently, the answer must include information that is uniquely attributable to the target paper within the given retrieval scope.
To achieve this, we utilize the Retrieval QA defined during the paper cluster construction stage.
Specifically, the Retrieval QA is reformulated so that its question explicitly highlights the target paper’s distinctive topic, variables, or findings, as in the following example:
\emph{“Which paper measured the incidence of pneumonia among Japanese participants in 2023?”}

\paragraph{Step 2. Generate Inter-document QA} Next, we construct a comparative or integrative question that explicitly links the Source and Target papers using only their combined QA pairs.
The question must reference shared variables, mechanisms, or outcomes appearing in the combined QA, avoiding abstract or generic formulations.
The answer is required to be logically entailed by the union of the two combined QA pairs and must not introduce external facts, unstated assumptions, or auxiliary evidence.

\paragraph{Step 3. Merge Complete QA} The results from the previous step are merged into a single, coherent multi-hop question that (i) requires locating the Target Paper and (ii) involves reasoning across both papers.
The LLM generates this merged question while preserving the intent of both components without introducing new information.
The final answer is either identical to, or deterministically derived from, the inter-document answer, ensuring consistency and a well-structured reasoning chain.

\paragraph{Step 4. QA Validation \emph{(Optional)}} Each Complete QA is evaluated on the criteria in Table~\ref{tab:mhqg_validation}.
Criterion is judged as either \textit{Accept} or \textit{Reject}; the \textit{Overall Decision} is determined from these binary ratings.
Items are retained only when the \textit{Overall Decision} is \textit{Accept}.
In the validation experiment, no items were rejected during the final evaluation phase.
To verify the functionality of the QA validation process, an additional analysis was conducted excluding the pre-rejected samples.
Results showed that approximately \textbf{70\% of pre-rejections originated from the \textit{Relational Appropriateness} or \textit{Cross-reference Necessity}} criteria, indicating that most filtering occurred when inter-paper connections were conceptually weak or unnecessary.
This confirms that the validation pipeline effectively distinguishes between coherent and incoherent inter-document QA constructions.

\begin{figure}[t]
    \centering
    \includegraphics[width=\columnwidth]{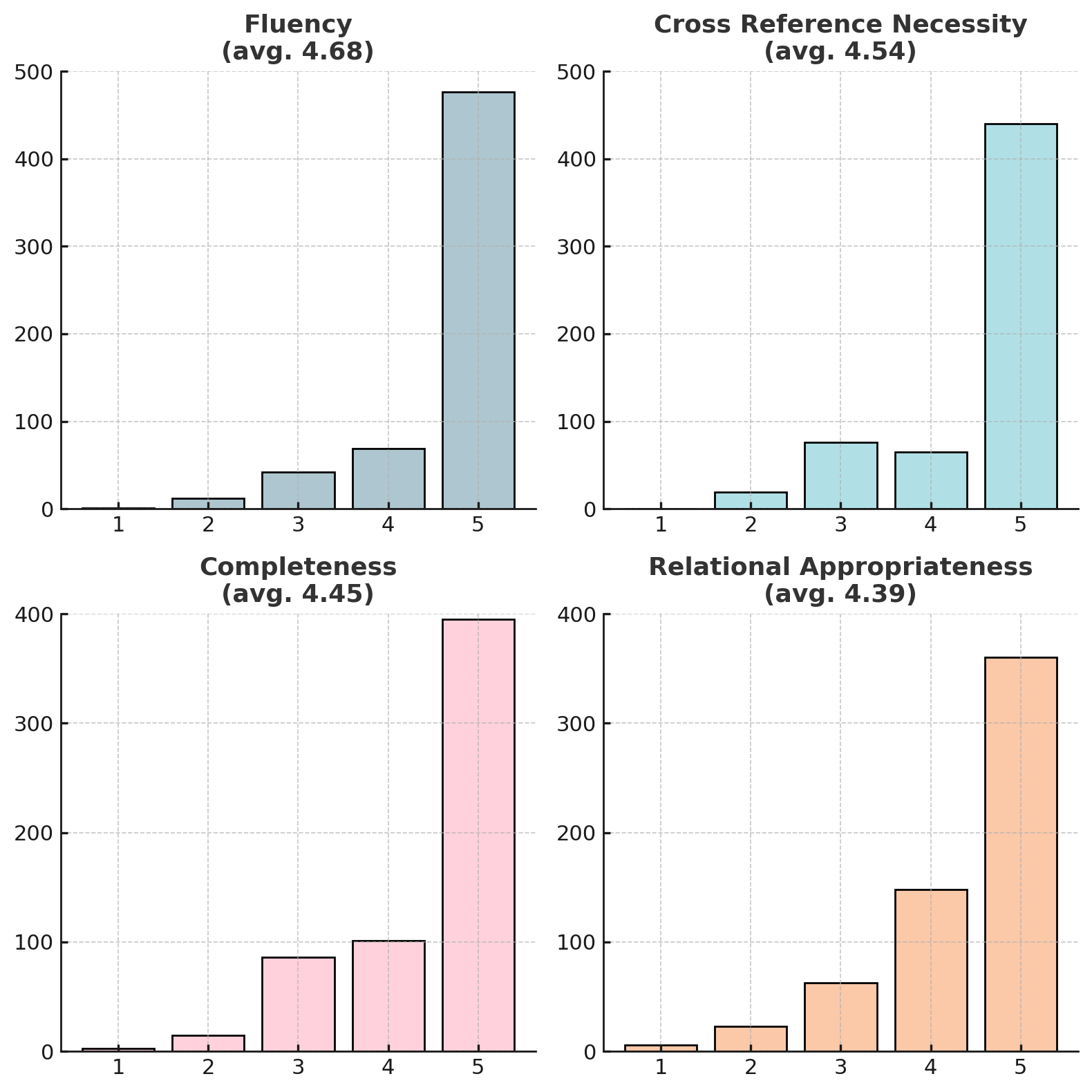}
    \caption{Results of multi-dimensional MHQA quality validation}
    \label{fig:quality_assurance_results}
\end{figure}

\paragraph{Quality Assurance}
Four graduate-level annotators independently rate 150 randomly sampled Complete QA pairs across four criteria as shown table \ref{tab:mhqg_validation} using a 5-point Likert scale.
Overall, the dataset showed strong linguistic and conceptual quality, with high mean scores and consistently coherent samples as shown in Figure \ref{fig:quality_assurance_results}.

\begin{table}[t]
\centering
\scriptsize
\setlength{\tabcolsep}{5pt}
\renewcommand{\arraystretch}{1.05}
\caption{Criteria of MHQA quality validation.}
\label{tab:mhqg_validation}
\resizebox{0.95\linewidth}{!}{%
\begin{tabular}{>{\centering\arraybackslash}m{0.32\linewidth}|>{\centering\arraybackslash}m{0.63\linewidth}}
\toprule
\textbf{Criterion} & \textbf{Description} \\
\midrule
\textbf{Fluency} & \vspace{0.5em}Whether the Complete QA is grammatically correct and naturally written.\vspace{0.5em} \\
\hline
\textbf{Completeness} & \vspace{0.3em}Whether the Complete QA can be answered solely based on the provided QA pair.\vspace{0.3em} \\
\hline
\textbf{Cross-reference Necessity} & \vspace{0.3em}Whether both papers are essential to answer the Complete QA.\vspace{0.3em} \\
\hline
\textbf{Relational Appropriateness} & \vspace{0.4em}Whether the two papers are conceptually suitable to be combined into a single QA.\vspace{0.4em} \\
\hline
\textbf{Overall Decision} & \vspace{0.3em}Synthesizes the above four criteria to make the final decision.\vspace{0.3em} \\
\bottomrule
\end{tabular}
}
\end{table}

\newcommand{\hlc}[2][yellow]{%
  \begingroup
  \setlength{\fboxsep}{1pt}%
  \colorbox{#1}{\strut #2}%
  \endgroup
}

\definecolor{LightPinkHL}{HTML}{FFC5D3}   
\definecolor{LightGreenHL}{HTML}{E8F6E8}  

\subsection{Verifications}
\label{sec:verifications}

\paragraph{Evidence Coverage Validation}
Beyond pair-level validation, we examined whether the SHQA list faithfully represents section content and whether retrieval questions remain uniquely specified within each paper cluster.
We first examined whether all salient information in a section was actually extracted as queries, since any unextracted information could undermine the uniqueness of the retrieval question.
To justify using SHQA as a surrogate for section information, we conducted a post-hoc audit of 40 sections containing 202 SHQA items that had passed manual validation and checked whether the evidence texts covered the meaningful parts of each section.
On average, 72\% of sentences in a section appeared as SHQA evidence, and this coverage rose to 88\% for sections such as abstract, result or conclusion, and discussion where independent facts are densely expressed.
When a single evidence span conveyed multiple individual facts, the content was consistently split into multiple SHQA items so that each item captured one fact per QA.
Notably, as shown in Table~\ref{tab:qa_examples_pmc10021650}, a single evidence passage often yields multiple question-answer pairs, resulting in more than 12,000 multi-QA evidence units. This design enriches contextual diversity and contributes to building a higher-quality SHQA dataset.

\paragraph{Retrieval Question Validation}
We then assessed the uniqueness of the retrieval question within the paper cluster and manually audited all test set retrieval questions.
These findings support the use of SHQA as a faithful surrogate for section content and establish that retrieval questions uniquely identify their intended targets under our clustering and question design constraints.

\begin{table}[h]
\caption{Example of Multi-QA Evidence}
\label{tab:qa_examples_pmc10021650}
\centering
\scriptsize
\setlength{\tabcolsep}{3pt}
\renewcommand{\arraystretch}{1.05}
\begin{tabularx}{0.98\linewidth}{@{}lX@{}}
\toprule
\textbf{Evidence (E)} &
For instance, %
{\color{RubineRed}the U.S. CDC N1 assay had a recall of 99.0\%} %
and %
{\color{ForestGreen}the U.S. CDC N2 assay had a 99.2\% recall.} \\
\textbf{Meta} & PMC10021650  \\
\midrule
\textbf{Q1} & What was the recall rate for the U.S. CDC N1 assay? \\
\textbf{A1} & The recall rate for the U.S. CDC N1 assay was 99.0\%. \\
\midrule
\textbf{Q2} & What was the recall rate for the U.S. CDC N2 assay? \\
\textbf{A2} & The recall rate for the U.S. CDC N2 assay was 99.2\%. \\
\bottomrule
\end{tabularx}
\end{table}
\section{Experiments}
The reliability and suitability of AIM-SciQA were validated in previous sections.
However, assessing the dataset against real-world task criteria represents a distinct challenge.
Therefore, to examine its practical applicability, we conduct a two-stage evaluation—retrieval and QA—on the IM-SciQA dataset. 

\begin{table*}[t]
\centering
\scriptsize
\caption{
Retrieval performance across different settings. 
\textbf{Bold} indicates the best performance, and \underline{underline} denotes the second best.
}
\label{tab:retrieval_all_sorted}
\resizebox{\textwidth}{!}{
\begin{tabular}{l|rrr|rrr|rrr}
\toprule
\multirow{2}{*}{\textbf{Model}} &
\multicolumn{3}{c|}{\textbf{Paper Cluster (N=30)}} &
\multicolumn{3}{c|}{\textbf{Random Cluster (N=30)}} &
\multicolumn{3}{c}{\textbf{Full Papers (N=8211)}} \\
\cmidrule(lr){2-10}
 & \textbf{Hit@1} & \textbf{Hit@3} & \textbf{MRR@5} 
 & \textbf{Hit@1} & \textbf{Hit@3} & \textbf{MRR@5} 
 & \textbf{Hit@1} & \textbf{Hit@50} & \textbf{MRR@50} \\
\midrule
GritLM-7B    \cite{muennighoff2024generative}     & 0.220 & 0.353 & 0.306 & 0.177 & 0.303 & 0.248 & 0.010 & 0.130 & 0.023 \\
Jina-v2-base-en \cite{gunther2023jina}           & 0.197 & 0.350 & 0.280 & 0.183 & 0.340 & 0.266 & 0.020 & 0.117 & 0.030 \\
e5-base-v2 \cite{wang2022text}              & 0.273 & 0.493 & 0.395 & 0.307 & 0.537 & 0.428 & 0.017 & 0.157 & 0.028 \\
e5-mistral-7b-instruct \cite{wang2023improving}    & 0.300 & 0.473 & 0.390 & 0.273 & 0.430 & 0.356 & 0.030 & 0.203 & 0.058 \\
Qwen3-Embedding-8B \cite{zhang2025qwen3}        & 0.407 & 0.597 & 0.554 & 0.483 & 0.607 & 0.555 & 0.080 & 0.397 & 0.136 \\
llm-embedder   \cite{xiao2024c}            & 0.497 & 0.683 & 0.666 & 0.577 & 0.753 & 0.665 & 0.130 & 0.480 & 0.190 \\
bge-large-en-v1.5  \cite{chen2024bge}        & 0.580 & 0.750 & 0.680 & 0.637 & 0.773 & 0.712 & \underline{0.213} & 0.550 & \underline{0.279} \\
text-embedding-ada-002 \cite{Greene2022openai}     & \underline{0.600} & \underline{0.783} & \underline{0.695} & \underline{0.667} & \underline{0.830} & \underline{0.752} & 0.207 & \underline{0.553} & 0.272 \\
voyage-3-large \cite{VoyageAI2025Voyage3Large}   & \textbf{0.673} & \textbf{0.830} & \textbf{0.761} & \textbf{0.737} & \textbf{0.887} & \textbf{0.811} & \textbf{0.240} & \textbf{0.637} & \textbf{0.325} \\
\bottomrule
\end{tabular}
}
\end{table*}

\subsection{Retrieval Evaluation}
We investigate the retrieval process for identifying the \textit{Target Paper} using diverse retrievers. 
As described in Section~\ref{sec:aim-sciqa}, each query in IM-SciQA is associated with a predefined \textit{paper cluster} constructed during the AIM-SciQA pipeline. 
A paper cluster is a compact set of 30 papers that includes one gold target and 29 distractor papers, derived either from citation-based or QA-based relational matching.
This cluster serves as a constrained retrieval environment that closely mirrors realistic scientific search scenarios, where relevant literature typically resides within a semantically or citation-linked subset of the corpus.

To analyze the impact of such clustering constraints, we design three experimental environments:
(1) \textit{Paper Cluster} — retrieval is performed within the original cluster defined by the AIM-SciQA construction process;
(2) \textit{Random Cluster} — the same number of documents (30) is randomly sampled per query to remove relational coherence; and
(3) \textit{w/o Cluster} — retrieval is conducted over the full corpus of 8,211 papers.
To ensure precise retrieval, we use the \textit{Retrieval Question} only, rather than the Combined QA, as the query input.

\paragraph{Models}
We evaluate various embedding models to analyze differences in representation quality and retrieval accuracy. 
All models were implemented based on the LlamaIndex framework \cite{Liu_LlamaIndex_2022}, and all experiments were conducted on an NVIDIA A100 (80GB) GPU.

\paragraph{Metrics}
To evaluate retrieval performance, we report \textbf{Hit@1, 3}, and \textbf{MRR@5} under the paper cluster setting ($N=30$), and \textbf{Hit@1, 50}, and \textbf{MRR@50} under the full paper setting ($N=8211$).
\textbf{Hit@K} measures the proportion of queries for which the correct document appears within the top K retrieved results, while \textbf{MRR@K (Mean Reciprocal Rank)} averages the reciprocal rank of the first correct document, reflecting the top-k ranking quality.

\subsection{IM-QA Evaluation}
We perform the \textbf{IM-QA} process using the given Source Paper and the retrieved Target Paper with various LLMs.
To disentangle the impact of the retrieval stage from the intrinsic IM-QA capability, we conduct evaluations under two settings: \textbf{Realistic} and \textbf{Oracle}. In the Realistic Setting, the top-1 document from the retrieval stage is provided as the Target Paper, whereas in the Oracle Setting, the gold Target Paper is supplied regardless of the retrieval outcome.

\paragraph{Models}
To obtain a representative view across model families, providers, and scales, we evaluate both open-source and proprietary LLMs. Inference for open-source models was performed using \textit{vLLM} \cite{kwon2023efficient} on an NVIDIA H100 GPU.
To further compare against human-level performance, two biomedical domain experts were asked to answer 100 QA items under the Oracle Setting.

\paragraph{Metrics}
As the main metric, we adopt LLM-based Accuracy, following {M3-SciQA}. Specifically, a strong LLM (GPT-5) is prompted with the Question, Answer, and Prediction to evaluate whether the Answer and Prediction provide the same response to the Question, scored on a 0–0.5–1 scale.
Traditional metrics such as token-level F1-score, ROUGE-L, and BERTScore \cite{he2021deberta} are also employed. 

\subsection{Results}

\begin{table*}[t]
\centering
\small
\caption{
Comparison of model performance under \textit{Oracle} and \textit{Realistic} settings.
\textbf{Bold} indicates the best performance, and \underline{underline} denotes the second best (based on Accuracy).
Accuracy in Oracle setting is measured over 100 QA instances.
}
\label{tab:qa_results}
\resizebox{\textwidth}{!}{
\begin{tabular}{l|cccc|cccc}
\toprule
\multirow{2}{*}{\textbf{Model}} &
\multicolumn{4}{c|}{\textbf{Oracle Setting}} &
\multicolumn{4}{c}{\textbf{Realistic Setting}} \\
\cmidrule(lr){2-9}
 & \textbf{Acc} & \textbf{F1} & \textbf{ROUGE} & \textbf{BERTScore} 
 & \textbf{Acc} & \textbf{F1} & \textbf{ROUGE} & \textbf{BERTScore} \\
\midrule
Human Expert & 0.975 & 0.324 & 0.319 & 0.447 & -- & -- & -- & -- \\
\midrule
gemma-3-1b-it \cite{team2025gemma} & 0.265 & 0.337 & 0.301 & 0.340 & 0.260 & 0.332 & 0.292 & 0.328 \\
Llama-3.2-3B-it \cite{grattafiori2024llama} & 0.57 & 0.421 & 0.403 & 0.435 & 0.515 & 0.399 & 0.380 & 0.403 \\
gemma-3-4b-it \cite{team2025gemma} & 0.590 & 0.430 & 0.403 & 0.465 & 0.515 & 0.418 & 0.390 & 0.453 \\
gemma-3-27b-it \cite{team2025gemma} & 0.750 & \underline{0.517} & \underline{0.495} & \underline{0.556} & 0.670 & \textbf{0.483} & \textbf{0.461} & \textbf{0.519} \\
Qwen3-30B-A3B-it \cite{yang2025qwen3} & 0.760 & 0.505 & 0.486 & 0.539 & 0.685 & 0.477 & 0.453 & 0.504 \\
DeepSeek-R1 \cite{guo2025deepseek} & 0.820 & 0.150 & 0.232 & 0.129 & 0.710 & 0.143 & 0.220 & 0.118 \\
gpt-oss-20b \cite{agarwal2025gpt} & 0.835 & 0.454 & 0.438 & 0.489 & 0.705 & 0.423 & 0.405 & 0.448 \\
claude-opus-4-1 \cite{Anthropic2025ClaudeOpus41} & \underline{0.855} & 0.476 & 0.464 & 0.522 & \underline{0.730} & 0.430 & 0.412 & 0.465 \\
gpt-5 \cite{OpenAI2025IntroducingGPT5} & \textbf{0.900} & \textbf{0.520} & \textbf{0.507} & \textbf{0.570} & \textbf{0.775} & \underline{0.474} & \underline{0.453} & \underline{0.512} \\
\bottomrule
\end{tabular}
}
\end{table*}

\paragraph{Retrieval Performance for Various Cluster Settings}

The results of our retrieval evaluation are presented in Table~\ref{tab:retrieval_all_sorted}.
Under the Paper Cluster setting, lower-capacity models showed performance comparable to the Random Cluster setting, whereas higher-capacity models exhibited significantly lower scores in both Hit@1,3.
A similar trend was observed in the Full Papers setting, where performance decreased despite the substantially larger search space (e.g., Hit@1 of 0.673 on 30 docs vs. 0.240 on 8,211 docs).
These results indicate that  \textbf{(1) the proposed Paper Clustering method effectively constructs a challenging and realistic candidate set that evaluates the fine-grained discriminative ability of retrieval models.}
Meanwhile, an error analysis of the Random Cluster setting revealed that many “incorrect” predictions actually contained information that could be regarded as valid alternative answers.
Together with the observation in Section~\ref{sec:verifications}, this finding supports that \textbf{(2) the cluster-level Unique Retrieval Question selection process effectively mitigates the risk of multiple valid answers.}

\paragraph{Performance Separation under Oracle Setting}

As shown in Table~\ref{tab:qa_results}, modern proprietary LLMs achieve near human-level performance under the \textit{Oracle Setting}, whereas smaller models such as \textit{Gemma-3-1B}, \textit{Gemma-3-4B}, and \textit{LLaMA-3B} exhibit notably lower accuracy even on relatively simple relational reasoning tasks.
The performance gap is more pronounced in the LLM-based Accuracy (GPT-Score) than in traditional metrics (F1, ROUGE-L, BERTScore), suggesting that smaller models tend to generate linguistically plausible yet factually inconsistent answers.
This pattern indicates that the required information can be effectively retrieved and reasoned over when the model has sufficient capacity, providing empirical evidence that both the question hints and the target information are well-grounded and meaningful.

\paragraph{Robustness and Sensitivity under Realistic Setting}

Meanwhile, as shown in Table~\ref{tab:qa_results}, all models experience a performance degradation under the \textit{Realistic Setting}, where the retrieved document is used as the target.
However, the degree of decline does not scale directly with the retrieval accuracy (Hit@1 = 0.673), implying that the reasoning process anchored on the Source Paper remains partially robust to retrieval noise.
Interestingly, stronger models that perform well under the Oracle Setting tend to show larger drops when retrieval is imperfect, suggesting that more capable models rely more heavily on precise evidence alignment.
This finding supports the view that providing the correct supporting document substantially facilitates inter-document reasoning and that the proposed dataset effectively evaluates such reasoning under realistic retrieval conditions.

\subsection{CIM-SciQA Evaluation}

\begin{table}[t]
\centering
\caption{CIM-SciQA Results (Realistic Setting)}
\label{tab:cim_sciqa_results}
\resizebox{0.48\textwidth}{!}{%
\begin{tabular}{lcccc}
\toprule
\textbf{Model} & \textbf{Acc} & \textbf{F1} & \textbf{ROUGE} & \textbf{BERTScore} \\
\midrule
gemma-3-1b-it \cite{team2025gemma} & 0.446 & 0.330 & 0.295 & 0.366 \\
gemma-3-4b-it \cite{team2025gemma} & 0.635 & 0.366 & 0.329 & 0.405 \\
Llama-3.2-3B-it \cite{grattafiori2024llama} & 0.608 & 0.367 & 0.339 & 0.379 \\
gemma-3-27b-it \cite{team2025gemma} & 0.798 & \underline{0.427} & \underline{0.402} & \underline{0.449} \\
DeepSeek-R1 \cite{guo2025deepseek} & 0.784 & 0.131 & 0.205 & 0.110 \\
Qwen3-30B-A3B-it \cite{yang2025qwen3} & 0.824 & 0.425 & 0.392 & 0.439 \\
gpt-oss-20b \cite{agarwal2025gpt} & \underline{0.851} & 0.373 & 0.346 & 0.381 \\
gpt-5 \cite{OpenAI2025IntroducingGPT5} & \textbf{0.926} & \textbf{0.432} & \textbf{0.411} & \textbf{0.478} \\
\bottomrule
\end{tabular}%
}
\end{table}

We evaluated \textbf{CIM-SciQA}, a citation-guided variant of AIM-SciQA, under the same evaluation configuration as the IM-SciQA dataset, except that we did not employ claude-opus-4-1 or human expert evaluation due to limited budget.
In this setting, the retrieval task of CIM-SciQA is conducted by locating appropriate in-text citation symbols within the source paper, while the realistic QA task is prompted to generate answers by referencing the full text of both the source paper and the paper corresponding to the cited symbol.
The evaluation results are presented in Table~\ref{tab:cim_sciqa_results}.

CIM-SciQA achieved performance comparable to the Oracle setting of IM-SciQA (Table~6), substantially surpassing the Realistic setting.
Even when considering the inherent complexity of the documents included in both datasets, this gap represents a significant difference.
We attribute this to the nature of citation-based questions, which are grounded in explicitly linked documents and therefore tend to form QA pairs of relatively lower difficulty.
\section{Conclusion}

We present \textbf{AIM-SciQA}, an automatic framework that constructs scientific IM-QA datasets, and introduce the resulting dataset, \textbf{IM-SciQA}.
Our method systematically combines SHQA extraction through both semantic and citation-based matching.
Using AIM-SciQA, we construct the IM-SciQA dataset from 8,211 PubMed Central articles, totaling 411k SHQA and 13K MHQA pairs with rigorous human and automatic validation.
Our experimental analyses show that recent LLMs remain inferior to human experts in factual consistency and multi-hop reasoning, when they exhibit strong linguistic fluency, and the gap is most pronounced under realistic retrieval settings.
These findings underscore the value of IM-SciQA as a challenging benchmark for evaluating retrieval-augmented reasoning and scientific comprehension in modern LLMs.
We hope AIM-SciQA provides a durable foundation for research on scalable scientific QA generation, improved retrieval–reasoning integration, and trustworthy multi-hop inference.
\section{Limitations}
Our current framework does not incorporate multimodal information such as figures or tables. This limitation arises due to lack of sufficiently powerful models capable of being effectively integrated into our automatic generation pipeline.
In addition, to ensure reliable multi-hop QA (MHQA) performance during automatic generation, we employed high-capacity large language models (LLMs) for combining multiple QA instances. While this approach improves answer consistency and reasoning quality, it incurs substantial computational and financial costs.
SHQA is designed to capture salient, answerable facts, not every detail of a paper. Consequently, it does not fully replicate the entire document. Still, the combination of strict exact-evidence matching, similarity-difference filtering, high inter-rater reliability (AC1 = 0.94), and section-level coverage (72–88\%) supports its adequacy as a pragmatic proxy for inter-document reasoning. Furthermore, our current work mainly focuses on inter-document reasoning, without exploring deeper multi-hop chains or diverse reasoning types. Expanding SHQA to include more complex reasoning hops and richer inference structures remains an important direction for future work.

\section{Acknowledgements}

This work was partly supported by the Institude of Information and Communications Technology Planning and Evaluation (IITP) grant funded by the Korean government (MSIT) (No. 2020-0-01361, Artificial Intelligence Graduate School Program (Yonsei University)).

\section{References}\label{sec:reference}
\bibliographystyle{lrec2026-natbib}
\bibliography{lrec2026-example}

@LanguageResource{park2025dochop,
  title={DocHop-QA: Towards Multi-Hop Reasoning over Multimodal Document Collections},
  author={Park, Jiwon and Pyeon, Seohyun and Kim, Jinwoo and Cabal, Rina Carines and Ding, Yihao and Han, Soyeon Caren},
  publisher={arXiv preprint arXiv:2508.15851},
  year={2025}
}

@LanguageResource{dasigi2021qasper,
  title={A Dataset of Information-Seeking Questions and Answers Anchored in Research Papers},
  author={Dasigi, Pradeep and Lo, Kyle and Beltagy, Iz and Cohan, Arman and Smith, Noah A and Gardner, Matt},
  publisher={Proceedings of the 2021 Conference of the North American Chapter of the Association for Computational Linguistics: Human Language Technologies},
  pages={4599--4610},
  year={2021}
}

@LanguageResource{lee2023qasa,
  title={QASA: Advanced Question Answering on Scientific Articles},
  author={Lee, Yukyung and others},
  booktitle={Proceedings of the 40th International Conference on Machine Learning},
  volume={202},
  pages={19036--19052},
  year={2023},
  organization={PMLR}
}

@LanguageResource{krithara2023bioasq,
  title={BioASQ-QA: A manually curated corpus for Biomedical Question Answering},
  author={Krithara, Anastasia and Nentidis, Anastasios and Bougiatiotis, Konstantinos and Paliouras, Georgios},
  publisher={Nature Scientific Data},
  volume={10},
  pages={2068},
  year={2023}
}

@article{pramanick2024spiqa,
  title={Spiqa: A dataset for multimodal question answering on scientific papers},
  author={Pramanick, Shraman and Chellappa, Rama and Venugopalan, Subhashini},
  journal={Advances in Neural Information Processing Systems},
  volume={37},
  pages={118807--118833},
  year={2024}
}

@article{li2024m3sciqa,
  title={M3SciQA: A multi-modal multi-document scientific QA benchmark for evaluating foundation models},
  author={Li, Chuhan and Shangguan, Ziyao and Zhao, Yilun and Li, Deyuan and Liu, Yixin and Cohan, Arman},
  journal={arXiv preprint arXiv:2411.04075},
  year={2024}
}

@article{jin2019pubmedqa,
  title={Pubmedqa: A dataset for biomedical research question answering},
  author={Jin, Qiao and Dhingra, Bhuwan and Liu, Zhengping and Cohen, William W and Lu, Xinghua},
  journal={arXiv preprint arXiv:1909.06146},
  year={2019}
}

@article{lu2024ai,
  title={The ai scientist: Towards fully automated open-ended scientific discovery},
  author={Lu, Chris and Lu, Cong and Lange, Robert Tjarko and Foerster, Jakob and Clune, Jeff and Ha, David},
  journal={arXiv preprint arXiv:2408.06292},
  year={2024}
}

@article{wan2024sciqag,
  title={SciQAG: A framework for auto-generated science question answering dataset with fine-grained evaluation},
  author={Wan, Yuwei and Liu, Yixuan and Ajith, Aswathy and Grazian, Clara and Hoex, Bram and Zhang, Wenjie and Kit, Chunyu and Xie, Tong and Foster, Ian},
  journal={arXiv preprint arXiv:2405.09939},
  year={2024}
}

@inproceedings{lee2023liquid,
  title={LIQUID: A framework for list question answering dataset generation},
  author={Lee, Seongyun and Kim, Hyunjae and Kang, Jaewoo},
  booktitle={Proceedings of the AAAI Conference on Artificial Intelligence},
  volume={37},
  pages={13014--13024},
  year={2023}
}

@article{lewis2020retrieval,
  title={Retrieval-augmented generation for knowledge-intensive nlp tasks},
  author={Lewis, Patrick and Perez, Ethan and Piktus, Aleksandra and Petroni, Fabio and Karpukhin, Vladimir and Goyal, Naman and K{\"u}ttler, Heinrich and Lewis, Mike and Yih, Wen-tau and Rockt{\"a}schel, Tim and others},
  journal={Advances in neural information processing systems},
  volume={33},
  pages={9459--9474},
  year={2020}
}

@inproceedings{moller2020covid,
  title={COVID-QA: A Question Answering Dataset for COVID-19},
  author={M{\"o}ller, Timo and Reina, Anthony and Jayakumar, Raghavan and Pietsch, Malte},
  booktitle={Proceedings of the 1st Workshop on NLP for COVID-19 at ACL 2020},
  pages={1--11},
  year={2020},
  organization={Association for Computational Linguistics}
}

@inproceedings{lu2022scienceqa,
  title={Learn to Explain: Multimodal Reasoning via Thought Chains for Science Question Answering},
  author={Lu, Peng and Bansal, Mohit and Xia, Tony and Liu, Siyuan and Sun, Shiyu and others},
  booktitle={Advances in Neural Information Processing Systems},
  pages={2--12},
  year={2022}
}

@inproceedings{koreeda2021contractnli,
  title={ContractNLI: A Dataset for Document-level Natural Language Inference for Contracts},
  author={Koreeda, Yuta and Manning, Christopher D},
  booktitle={Findings of the Association for Computational Linguistics: EMNLP 2021},
  pages={1907--1919},
  year={2021}
}

@article{amiri2025chemrxivquest,
  title={ChemRxivQuest: A Curated Chemistry Question-Answer Database Extracted from ChemRxiv Preprints},
  author={Amiri, Mahmoud and Bocklitz, Thomas},
  journal={arXiv preprint arXiv:2505.05232},
  year={2025}
}

@inproceedings{baumgartner2025peerqa,
  title={PeerQA: A Scientific Question Answering Dataset from Peer Reviews},
  author={Baum\"{g}rtner, Tim and Briscoe, Ted and Gurevych, Iryna},
  booktitle={Proceedings of the 2025 Conference of the North American Chapter of the Association for Computational Linguistics: Human Language Technologies (Volume 1: Long Papers)},
  pages={508--544},
  year={2025}
}

@inproceedings{hwang2024e2eqr,
  title={Explainable Multi-hop Question Generation: An End-to-End Approach without Intermediate Question Labeling},
  author={Hwang, Seonjeong and Kim, Yunsu and Lee, Gary Geunbae},
  booktitle={Proceedings of the 2024 Joint International Conference on Computational Linguistics, Language Resources and Evaluation (LREC-COLING 2024)},
  pages={6855--6866},
  year={2024}
}

@inproceedings{lin2024tase,
  title={Prompting Few-shot Multi-hop Question Generation via Comprehending Type-aware Semantics},
  author={Lin, Zefeng and Chen, Weidong and Song, Yan and Zhang, Yongdong},
  booktitle={Findings of the Association for Computational Linguistics: NAACL 2024},
  pages={3730--3740},
  year={2024}
}

@article{team2025gemma,
  title={Gemma 3 technical report},
  author={Team, Gemma and Kamath, Aishwarya and Ferret, Johan and Pathak, Shreya and Vieillard, Nino and Merhej, Ramona and Perrin, Sarah and Matejovicova, Tatiana and Ram{\'e}, Alexandre and Rivi{\`e}re, Morgane and others},
  journal={arXiv preprint arXiv:2503.19786},
  year={2025}
}

@article{yang2025qwen3,
  title={Qwen3 technical report},
  author={Yang, An and Li, Anfeng and Yang, Baosong and Zhang, Beichen and Hui, Binyuan and Zheng, Bo and Yu, Bowen and Gao, Chang and Huang, Chengen and Lv, Chenxu and others},
  journal={arXiv preprint arXiv:2505.09388},
  year={2025}
}

@article{grattafiori2024llama,
  title={The llama 3 herd of models},
  author={Grattafiori, Aaron and Dubey, Abhimanyu and Jauhri, Abhinav and Pandey, Abhinav and Kadian, Abhishek and Al-Dahle, Ahmad and Letman, Aiesha and Mathur, Akhil and Schelten, Alan and Vaughan, Alex and others},
  journal={arXiv preprint arXiv:2407.21783},
  year={2024}
}

@article{agarwal2025gpt,
  title={gpt-oss-120b \& gpt-oss-20b model card},
  author={Agarwal, Sandhini and Ahmad, Lama and Ai, Jason and Altman, Sam and Applebaum, Andy and Arbus, Edwin and Arora, Rahul K and Bai, Yu and Baker, Bowen and Bao, Haiming and others},
  journal={arXiv preprint arXiv:2508.10925},
  year={2025}
}

@inproceedings{kwon2023efficient,
  title={Efficient memory management for large language model serving with pagedattention},
  author={Kwon, Woosuk and Li, Zhuohan and Zhuang, Siyuan and Sheng, Ying and Zheng, Lianmin and Yu, Cody Hao and Gonzalez, Joseph and Zhang, Hao and Stoica, Ion},
  booktitle={Proceedings of the 29th symposium on operating systems principles},
  pages={611--626},
  year={2023}
}

@article{chen2024bge,
  title={Bge m3-embedding: Multi-lingual, multi-functionality, multi-granularity text embeddings through self-knowledge distillation},
  author={Chen, Jianlv and Xiao, Shitao and Zhang, Peitian and Luo, Kun and Lian, Defu and Liu, Zheng},
  journal={arXiv preprint arXiv:2402.03216},
  year={2024}
}

@article{wang2022text,
  title={Text embeddings by weakly-supervised contrastive pre-training},
  author={Wang, Liang and Yang, Nan and Huang, Xiaolong and Jiao, Binxing and Yang, Linjun and Jiang, Daxin and Majumder, Rangan and Wei, Furu},
  journal={arXiv preprint arXiv:2212.03533},
  year={2022}
}

@article{gunther2023jina,
  title={Jina embeddings 2: 8192-token general-purpose text embeddings for long documents},
  author={G{\"u}nther, Michael and Ong, Jackmin and Mohr, Isabelle and Abdessalem, Alaeddine and Abel, Tanguy and Akram, Mohammad Kalim and Guzman, Susana and Mastrapas, Georgios and Sturua, Saba and Wang, Bo and others},
  journal={arXiv preprint arXiv:2310.19923},
  year={2023}
}

@article{zhang2025qwen3,
  title={Qwen3 Embedding: Advancing Text Embedding and Reranking Through Foundation Models},
  author={Zhang, Yanzhao and Li, Mingxin and Long, Dingkun and Zhang, Xin and Lin, Huan and Yang, Baosong and Xie, Pengjun and Yang, An and Liu, Dayiheng and Lin, Junyang and others},
  journal={arXiv preprint arXiv:2506.05176},
  year={2025}
}

@inproceedings{muennighoff2024generative,
  title={Generative representational instruction tuning},
  author={Muennighoff, Niklas and Hongjin, SU and Wang, Liang and Yang, Nan and Wei, Furu and Yu, Tao and Singh, Amanpreet and Kiela, Douwe},
  booktitle={The Thirteenth International Conference on Learning Representations},
  year={2024}
}

@article{gwet2008computing,
  title={Computing inter-rater reliability and its variance in the presence of high agreement},
  author={Gwet, Kilem Li},
  journal={British Journal of Mathematical and Statistical Psychology},
  volume={61},
  number={1},
  pages={29--48},
  year={2008},
  publisher={Wiley Online Library}
}

@article{landis1977measurement,
  title={The measurement of observer agreement for categorical data},
  author={Landis, J Richard and Koch, Gary G},
  journal={biometrics},
  pages={159--174},
  year={1977},
  publisher={JSTOR}
}

@article{guo2025deepseek,
  title={Deepseek-r1: Incentivizing reasoning capability in llms via reinforcement learning},
  author={Guo, Daya and Yang, Dejian and Zhang, Haowei and Song, Junxiao and Zhang, Ruoyu and Xu, Runxin and Zhu, Qihao and Ma, Shirong and Wang, Peiyi and Bi, Xiao and others},
  journal={arXiv preprint arXiv:2501.12948},
  year={2025}
}

@article{gottweis2025towards,
  title={Towards an AI co-scientist},
  author={Gottweis, Juraj and Weng, Wei-Hung and Daryin, Alexander and Tu, Tao and Palepu, Anil and Sirkovic, Petar and Myaskovsky, Artiom and Weissenberger, Felix and Rong, Keran and Tanno, Ryutaro and others},
  journal={arXiv preprint arXiv:2502.18864},
  year={2025}
}

@article{wang2023improving,
  title={Improving text embeddings with large language models},
  author={Wang, Liang and Yang, Nan and Huang, Xiaolong and Yang, Linjun and Majumder, Rangan and Wei, Furu},
  journal={arXiv preprint arXiv:2401.00368},
  year={2023}
}

@misc{qwen2.5,
    title = {Qwen2.5: A Party of Foundation Models},
    url = {https://qwenlm.github.io/blog/qwen2.5/},
    author = {Qwen Team},
    month = {September},
    year = {2024}
}

@misc{balachandran2024medembed,
  author = {Balachandran, Abhinand},
  title = {MedEmbed: Medical-Focused Embedding Models},
  year = {2024},
  url = {https://github.com/abhinand5/MedEmbed}
}

@article{liu2023lost,
  title={Lost in the middle: How language models use long contexts},
  author={Liu, Nelson F and Lin, Kevin and Hewitt, John and Paranjape, Ashwin and Bevilacqua, Michele and Petroni, Fabio and Liang, Percy},
  journal={arXiv preprint arXiv:2307.03172},
  year={2023}
}

@inproceedings{xiao2024c,
  title={C-pack: Packed resources for general chinese embeddings},
  author={Xiao, Shitao and Liu, Zheng and Zhang, Peitian and Muennighoff, Niklas and Lian, Defu and Nie, Jian-Yun},
  booktitle={Proceedings of the 47th international ACM SIGIR conference on research and development in information retrieval},
  pages={641--649},
  year={2024}
}

@article{wadden2022scifact,
  title={SciFact-open: Towards open-domain scientific claim verification},
  author={Wadden, David and Lo, Kyle and Kuehl, Bailey and Cohan, Arman and Beltagy, Iz and Wang, Lucy Lu and Hajishirzi, Hannaneh},
  journal={arXiv preprint arXiv:2210.13777},
  year={2022}
}

@article{shi2024generate,
  title={Generate-then-ground in retrieval-augmented generation for multi-hop question answering},
  author={Shi, Zhengliang and Sun, Weiwei and Gao, Shen and Ren, Pengjie and Chen, Zhumin and Ren, Zhaochun},
  journal={arXiv preprint arXiv:2406.14891},
  year={2024}
}

@inproceedings{
he2021deberta,
    title={DEBERTA: DECODING-ENHANCED BERT WITH DISENTANGLED ATTENTION},
    author={Pengcheng He and Xiaodong Liu and Jianfeng Gao and Weizhu Chen},
    booktitle={International Conference on Learning Representations},
    year={2021},
    url={https://openreview.net/forum?id=XPZIaotutsD}
}

@misc{Anthropic2025ClaudeOpus41,
  author       = {Anthropic},
  title        = {Claude Opus 4.1},
  howpublished = {https://www.anthropic.com/news/claude-opus-4-1},
  month        = aug,
  year         = {2025},
  note         = {News release},
}

@misc{Greene2022openai,
  author       = {Ryan Greene and Ted Sanders and Lilian Weng and Arvind Neelakantan},
  title        = {New and Improved Embedding Model},
  howpublished = {https://openai.com/index/new-and-improved-embedding-model},
  month        = dec,
  year         = {2022},
  note         = {OpenAI Blog},
}

@misc{VoyageAI2025Voyage3Large,
  author       = {VoyageAI},
  title        = {voyage-3-large: the new state-of-the-art general-purpose and multilingual embedding model that ranks first across eight evaluated domains spanning 100 datasets},
  howpublished = {https://blog.voyageai.com/2025/01/07/voyage-3-large},
  month        = jan,
  day          = 7,
  year         = 2025,
  note         = {Blog post},
}

@misc{Liu_LlamaIndex_2022,
author = {Liu, Jerry},
doi = {10.5281/zenodo.1234},
month = {11},
title = {{LlamaIndex}},
url = {https://github.com/jerryjliu/llama_index},
year = {2022}
}

@misc{OpenAI2025IntroducingGPT5,
  author       = {OpenAI},
  title        = {Introducing GPT-5},
  howpublished = {https://openai.com/index/introducing-gpt-5},
  month        = aug,
  day          = 7,
  year         = {2025},
  note         = {Blog post},
}

\appendix
\newpage
\onecolumn

\section{Prompts}

\subsection{Single-hop QA Generation Prompt}
\label{app:shqa_prompt}

\begin{verbatim}
You are an expert in reading comprehension.
You will be provided with a single paragraph from a scientific paper.
Read the paragraph carefully and identify meaningful information
(entities, facts, relations, events) that can be directly and
unambiguously extracted from it.

For each identified meaningful fact, you must generate a JSON object
with the following three fields:

question: Generate one clear and specific question. The question must
detail the context or conditions (e.g., the specific patient group,
timeframe, or situation) mentioned in the text to pinpoint the fact
accurately.

evidence: Extract the minimal contiguous text span from the paragraph
that provides the evidence for the answer. This must be an exact quote
from the text.

answer: Based on the extracted evidence, formulate a natural complete
sentence that directly answers the question.

Rules:

Do not use any outside knowledge or inference.
All information must be sourced only from the given paragraph.

The evidence field must be an exact extraction without modification.

Avoid generating reasoning-type (multi-hop) questions.

Return the result as a JSON list with the following structure:

{
"question": "string",
"evidence": "string",
"answer": "string"
}
\end{verbatim}

\subsection{Multi-hop QA Generation Prompt}
\label{app:mhqa_prompt}

\subsubsection{Reasoning Prompt}

\begin{verbatim}
You are a scientific QA generator for inter-document multi-hop
question construction.

Your task is to produce a structured QA reasoning process that
constructs an inter-document multi-hop question across multiple
scientific papers.

Step 0. Validation (not included in output)

Before generating any reasoning output, verify whether the Source QA
and Target QA form a meaningful multi-hop relationship such as
comparison, causation, conceptual linkage, or inference.

If no logical or conceptual overlap exists, output only:

Cannot generate inter-document multi-hop question.

Reasoning Components

Find Target Paper

Transform the Unique Target QA question into a question that asks
which paper addresses that topic.

The answer must identify the target paper using the format:

Identified target paper: '[Target Paper Title]'.

Generate Inter-document QA

Construct a comparative or integrative question connecting the
Source Paper and Target Paper using the Core QA pair.

The answer must be logically derived only from the provided QA pair.

Merge Complete QA

Combine the Find Target Paper question and the Inter-document QA
question into a single multi-hop question.

The final answer must be identical to or directly derived from the
Inter-document QA answer.

QA Validation

Evaluate the generated QA using the following criteria:

Fluency
Completeness
Cross-reference Necessity
Relational Appropriateness

Each criterion is labeled Accept or Reject and followed by an
overall Decision.

Output Format

<outputs>
  <component type="Find Target Paper">
    <question>...</question>
    <answer>...</answer>
  </component>

  <component type="Generate Inter-document QA">
    <question>...</question>
    <answer>...</answer>
  </component>

  <component type="Merge Complete QA">
    <question>...</question>
    <answer>...</answer>
  </component>

  <component type="QA Validation">
    <score type="Fluency">...</score>
    <score type="Completeness">...</score>
    <score type="Cross-reference Necessity">...</score>
    <score type="Relational Appropriateness">...</score>
    <score type="Decision">...</score>
  </component>
</outputs>
\end{verbatim}

\subsubsection{Input Template}

\begin{verbatim}
### Now Your Turn

<inputs>
  <source_paper>
    <title>{source_paper_title}</title>
    <section_name>{source_section_name}</section_name>
    <core_qa>
      <question>{core_source_q}</question>
      <answer>{core_source_a}</answer>
    </core_qa>
  </source_paper>
  <target_paper>
    <title>{target_paper_title}</title>
    <section_name>{target_section_name}</section_name>
    <core_qa>
      <question>{core_target_q}</question>
      <answer>{core_target_a}</answer>
    </core_qa>
    <unique_qa>
      <question>{unique_target_q}</question>
      <answer>{unique_target_a}</answer>
    </unique_qa>
  </target_paper>
</inputs>
\end{verbatim}

\end{document}